\documentclass{elsart}
\usepackage{algorithm,algorithmic,amsmath,amssymb,framed,graphicx,multirow}
\usepackage{graphicx,amssymb}
\usepackage{algorithmic}
\usepackage{algorithm}
\usepackage{amsmath}

\usepackage{appendix}
\usepackage{longtable}
\usepackage{booktabs}

\usepackage[switch,pagewise]{lineno}
\usepackage[usenames]{color}
\usepackage{url}

\emergencystretch=\maxdimen
\begin{document}
\begin{frontmatter}

\title{Iterated two-phase local search for the Set-Union Knapsack Problem}
\author[Angers]{Zequn Wei}
\and
\author[Angers,IUF]{Jin-Kao Hao\corauthref{cor}}
\corauth[cor]{Corresponding author.} \ead{jin-kao.hao@univ-angers.fr}
\address[Angers]{LERIA, Universit$\acute{e}$ d'Angers, 2 Boulevard Lavoisier, 49045 Angers, France}
\address[IUF]{Institut Universitaire de France, 1 Rue Descartes, 75231  Paris, France}

\maketitle

\begin{abstract}
The Set-union Knapsack Problem (SUKP) is a generalization of the popular 0-1 knapsack problem. Given a set of weighted elements and a set of items with profits  where each item is composed of a subset of elements, the SUKP involves packing a subset of items in a capacity-constrained knapsack such that the total profit of the selected items is maximized while their weights do not exceed the knapsack capacity. In this work, we present an effective iterated two-phase local search algorithm for this NP-hard combinatorial optimization problem. The proposed algorithm iterates through two search phases: a local optima exploration phase that alternates between a variable neighborhood descent search and a tabu search to explore local optimal solutions, and a local optima escaping phase to drive the search to unexplored regions. We show the competitiveness of the algorithm compared to the state-of-the-art methods in the literature. Specifically, the algorithm discovers 18 improved best results (new lower bounds) for the 30 benchmark instances and matches the best-known results for the 12 remaining instances. We also report the first computational results with the general CPLEX solver, including 6 proven optimal solutions. Finally, we investigate the effectiveness of the key ingredients of the algorithm on its performance.

\emph{Keywords}: Knapsack problems; Computational methods; Heuristics and metaheuristics; Combinatorial optimization.
\end{abstract}

\end{frontmatter}

\section{Introduction}
\label{Sec_Intro}
Given $U =  \{1,\ldots, n\}$ be a set of $n$ elements where each element $j \ (j = 1, \ldots , n)$ has a weight $w_j > 0$, we consider a set of $m$ items $V =  \{1,\ldots, m\}$  where each item $i \ (i = 1, \ldots , m)$ corresponds to a subset of elements $U_i \subset U$ determined by a relation matrix and has a profit $p_i > 0$. For an arbitrary non-empty item set $S \subset V$, the total profit of $S$ is defined as $P(S) = \sum_{i \in S}p_i$, and the weight of $S$ is given by $W(S) = \sum_{j \in \cup _{i \in S} U_i} w_j$. Let $C>0$ be the capacity of a given knapsack, the SUKP involves finding a subset of items $S^* \subset V$ such that the profit $P(S^*)$ is maximized and the weight $W(S^*)$ does not surpass the knapsack capacity $C$. Formally, the SUKP can be stated as follows.

\begin{equation}\label{MAX}
(SUKP) \quad \quad \mathrm{Maximize} \quad P(S) = \sum\limits_{i \in S} p_i
\end{equation}
\begin{equation}\label{constraint1}
\mathrm{s.t.} \quad  W(S) = \sum\limits_{j \in \cup_{i \in S} U_i} w_j \leq C, \ S \subset V
\end{equation}	
	
It is worth noting that for a given subset $S$ of items, the weight $w_j$ of an element $j$ is counted only once in $W(S)$ even if the element belongs to more than one selected items. 

One notices that the popular NP-hard 0-1 knapsack problem (KP) \cite{kellerer2003knapsack} is a special case of the SUKP. Indeed, the SUKP reduces to the KP when we set $m=n$ and $V=U$. The SUKP also generalizes the NP-hard densest k-subhypergraph problem (DkSH) that aims to determine a set of $k$ nodes of a hypergraph to maximize the number of hyperedges of the subhypergraph induced by the set of the selected nodes \cite{Chlamtacetal2018}. In fact, the SUKP reduces to the DkSH when we consider the elements and items as the nodes and hyperedges of a hypergraph respectively, with unit weights and unit profits as well as a capacity of $k$. As indicated in \cite{goldschmidt1994note,he2018novel}, the SUKP has a number of relevant applications, such as financial decision making, flexible manufacturing, building public key prototype, database partitioning etc. However, as a generalization of the NP-hard KP and DkSH problems, the SUKP is computationally challenging.

Given its theoretical and practical significance, the SUKP has received more and more attention. For instance, in 1994, Goldschmidt et al. devised an exact method for the SUKP based on the dynamic programming method \cite{goldschmidt1994note}. In 2014, Arulselvan studied a greedy approximation algorithm based on an approximation algorithm for the related budgeted maximum coverage problem \cite{arulselvan2014note}. In 2016, Taylor designed an approximation algorithm using results of the related densest k-subhypergraph problem \cite{taylor2016approximations}. 

In addition to the above exact and approximation algorithms, metaheuristic algorithms based on swarm optimization were recently studied to find sub-optimal solutions for the SUKP \cite{fengetal2019,he2018novel,ozsoydan2018swarm}. In 2017, He et al. proposed the first binary artificial bee colony algorithm (BABC) for sovling the SUKP and presented large scale experiments on a set of 30 SUKP benchmark instances \cite{he2018novel}. The comparisons with three other population-based algorithms show a competitive performance of BABC. In 2018, Ozsoydan and Baykasoglu presented a binary particle swarm optimization algorithm (gPSO) \cite{ozsoydan2018swarm} and reported improved best results on the set of 30 benchmark instances of \cite{he2018novel}. In 2019, Feng et al. investigated several versions of discrete moth search (MS) and tested on 15 out of the 30 SUKP benchmark instances \cite{fengetal2019} and also presented some updated best results. In this work, we will use these recent algorithms as the main references for our computational studies. 

We observe that the state-of-the-art algorithms of \cite{fengetal2019,he2018novel,ozsoydan2018swarm} adopted swam optimization methods that are initially designed for solving continuous problems. Given that the SUKP is a discrete (binary) problem, these swam algorithms integrate various adaptations to cope with the binary feature of the problem. As such, they simulate the discrete optimization via continuous search operators and strategies. In this work, we investigate for the first time stochastic local search for solving the SUKP, which directly operates in the binary search space. This work is motivated by two considerations. First, stochastic local search has been quite successful in solving numerous combinatorial problems \cite{Hoos2004}. Second, for many knapsack problems, the best performing algorithms are based on local optimization approaches; e.g., multidimensional knapsack problem \cite{glover1996critical,lai2018two2,vasquez2001hybrid}, multidemand multidimensional knapsack problem \cite{cappanera2005local,lai2018two}, multiple-choice multidimensional knapsack problem \cite{ChenHao2014,hifi2006reactive}, quadratic knapsack problem \cite{chen2017iterated,yang2013effective}, quadratic multiple knapsack problem \cite{ChenHao2015,peng2016ejection} and generalized quadratic knapsack problem \cite{avci2017multi}. In this work, we show that the discrete optimization approach based on stochastic local search is also quite valuable and effective for solving the SUKP.

The contributions of this work are summarized as follows.

From a perspective of algorithm design, we present the iterated two-phase local search algorithm (I2PLS) that integrates an intensification-oriented component (first phase) and a diversification-oriented component (second phase). The first phase combines a descent search procedure and a tabu search procedure to explore various local optimal solutions based on dedicated neighborhoods. The second phase diversifies the search by performing a frequency-guided perturbation. 

From a perspective of computational performance, we show the competitiveness of the proposed algorithm compared to the state-of-the-art algorithms on the set of 30 benchmark instances commonly used in the literature. In particular, we report improved best results for 18 large instances and equal best results for the 12 remaining instances. The improved best results (new lower bounds) are useful for future studies on the problem, e.g., they can serve as references for evaluating existing and new SUKP algorithms.

Third, we investigate for the first time the interest of the general mixed integer programming solver CPLEX for solving the SUKP. We show that while CPLEX (version 12.8) can find the optimal solutions for the 6 small benchmark instances (with 85 to 100 items and elements) based on a simple 0/1 linear programming model, it fails to exactly solve the other 24 instances. These outcomes provide another motivation for developing effective approximate algorithms to handle problem instances that cannot be solved exactly.

The remaining part of this paper is organized as follows. In Section \ref{Sec_Approach}, we present the general framework of the proposed algorithm as well as its composing ingredients. Computational results and comparisons with the best-performing algorithms and CPLEX are reported in Section 3. In Section 4, we analyze the parameters and components of the algorithm and show their effects on its performance. Finally, we summarize the present work and discuss future research directions.

\section{Iterated two-phase local search for the SUKP}
\label{Sec_Approach}

This section is dedicated to the presentation of the proposed iterated two-phase local search algorithm (I2PLS) for the SUKP. We first show its general scheme, and then explain the composing ingredients.

\subsection{General Algorithm}
\label{General Algorithm}

As shown in Algorithm \ref{Algo_I2PLS}, I2PLS is composed of two complementary search phases: a local optima exploration phase (Explore) to find new local optimal solutions of increasing quality and a local optima escaping phase (Escape) to displace the search to unexplored regions.

\begin{algorithm}
\footnotesize
\caption{Iterated two-phase local search for the SUKP}\label{Algo_I2PLS}
\begin{algorithmic}[1]
 \STATE \textbf{Input}: Instance $I$, cut-off time $t_{max}$, neighborhoods $N_1-N_3$, exploration depth $\lambda_{max}$, sampling probability $\rho$, tabu search depth $\omega_{max}$, perturbation strength $\eta$.
   \STATE \textbf{Output}: The best solution found $S^*$.
   \STATE /* Generate an initial solution $S_0$ in a greedy way, \S \ref{Initialization} \hfill */
   \\ $S_0 \gets  Greedy\_Initial\_Solution(I) $
   \STATE $S^* \gets S_0$ \hfill /* Record the overall best solution $S^*$ found so far */ 
   \WHILE{ $Time \leq t_{max}$}
   \STATE /* Local optima exploration phase using VND and TS, \S \ref{Exploration} \hfill */
   \\ $S_b \gets $VND-TS$(S_0, N_1-N_3, \lambda_{max}, \rho, \omega_{max}) $
   \IF{$f(S_b) > f(S^*)$}
   \STATE $S^* \gets S_b$ \hfill /* Update the best solution $S^*$ found so far */ 
   \ENDIF
   \STATE /* Local optima escaping phase using frequency-based perturbation, \S \ref{Escaping} \hfill */
   \\$S_0 \gets $Frequency\_Based\_Local\_Optima\_Escaping$(S_b,\eta)$
   \ENDWHILE   
   \RETURN $S^*$
\end{algorithmic}
\end{algorithm}

The algorithm starts from a feasible initial solution (line 3, Alg. \ref{Algo_I2PLS}) that is obtained with a greedy construction procedure (Section \ref{Initialization}). Then it enters the `while' loop to iterate the `Explore' phase and the `Escape' phase (lines 5-11, Alg. \ref{Algo_I2PLS}) to seek solutions of improving quality. At each iteration, the `Explore' phase (line 6, Alg. \ref{Algo_I2PLS}) first performs a variable neighborhood descent (VND) search to locate a new local optimal solution within two neighborhoods $N_1$ and $N_2$ and then runs a tabu search (TS) to explore additional local optima with a different neighborhood $N_3$ (Section \ref{Exploration}). When the `Explore' phase is exhausted, I2PLS switches to the `Escape' phase  (line 10, Alg. \ref{Algo_I2PLS}), which uses a frequency-based perturbation to displace the search to an unexplored region (Section \ref{Escaping}). These two phases are iterated until a stopping condition (in our case, a given time limit $t_{max}$) is reached. During the search process, the best solution found is recorded in $S^*$ (lines 7-8, Alg. \ref{Algo_I2PLS}) and returned as the final output of the algorithm at the end of the algorithm. 

One notices that the general scheme of the I2PLS algorithm for the SUKP shares ideas of breakout local search \cite{benlic2013breakout1}, three-phase local search \cite{fu2015three} and iterated local search \cite{lourencco2003iterated}. Meanwhile, to ensure its effectiveness for solving the SUKP, the proposed algorithm integrates dedicated search components tailored for the considered problem, which are described below.

\subsection{Solution Representation, Search Space, and Evaluation Function}
\label{subsec_space}

Given a SUKP instance composed of $m$ items $V =  \{1, \ldots , m\}$, $n$ elements $U =  \{1, \ldots , n\}$ and knapsack capacity $C$. The search space $\Omega$ includes all non-empty subsets of items such that the capacity constraint is satisfied.

\begin{equation}\label{Search Space}
\Omega = \{S \subset V: S \neq  \emptyset, \sum\limits_{j \in \cup_{i \in S} U_i} w_j \leq C\}
\end{equation}

For any candidate solution $S$ of $\Omega$, its quality is assessed by the objective value $f(S)$ that corresponds to the total profit of the selected items, 

\begin{equation}\label{Eva}
f(S) = \sum\limits_{i \in S} p_i
\end{equation}

Notice that a candidate solution $S$ of $\Omega$ can be represented by $S =<A,\bar{A}>$ where $A$ is the set of selected items and $\bar{A}$ are the non-selected items. 

The goal of our I2PLS algorithm is to find a solution $S \in \Omega $ with the objective value $f(S)$ as large as possible. 

\subsection{Initialization}
\label{Initialization}

The I2PLS algorithm starts its search with an initial solution, which is generated by a simple greedy procedure in three steps. First, we calculate the total weight $w_i$ of each item $i$ in $O(mn)$. Second, based on the given profit $p_i$ of each item, we obtain the \textit{profit ratio} $r_i$ of each item by $r_i = p_i / w_i$ and sort all items in the descending order according to $r_i$ in $O(log(m))$. Third, we add one by one the items to $S$ by following this order until the capacity of the knapsack is reached in $O(m)$. The time complexity of the initialization procedure is thus $O(mn)$. 

\subsection{Local Optima Exploration Phase}
\label{Exploration}

\begin{algorithm}
\footnotesize
\caption{Local Optima Exploration Phase - VND-TS}\label{Algo_LOExplore}
\begin{algorithmic}[1]
   \STATE \textbf{Input}: Starting solution $S$, neighborhoods $N_1-N_3$, exploration depth $\lambda_{max}$, sampling probability $\rho$, tabu search depth $\omega_{max}$, 
   \STATE \textbf{Output}: The best solution $S_b$ found by VND-TS.
   \STATE $S_b \gets S$     \hfill	/*$S_b$ records the best solution found so far during VND-TS */
   \STATE $\lambda \gets 0$   \hfill	/*$\lambda$ counts the number of consecutive non-improving rounds*/
   \WHILE {$\lambda < \lambda_{max}$}
   \STATE /* Attain a new local optimum $S$ by VND with $N_1$ and $N_2$, see Alg. \ref{Algo_Descent} \hfill */ \\
    	  $S \gets $VND$(S,N_1,N_2,\rho) $ 
   \STATE /* Explore nearby optima around the new $S$ by TS with $N_3$, see Alg. \ref{Algo_Tabu} \hfill  */   
   \\     $(S_c,S) \gets $TS$(S,N_3,\omega_{max}) $ /*$S_c$ is the best solution found so far during TS */
   \IF    {$f(S_c) > f(S_b)$}
   \STATE $S_b \gets S_c$ \hfill /* Update the best solution $S_b$ found so far */      
   \STATE $\lambda \gets 0$
   \ELSE 
   \STATE $\lambda \gets \lambda + 1$
   \ENDIF 
   \ENDWHILE
   \RETURN $S_b$
\end{algorithmic}
\end{algorithm}

From an initial solution, the `Explore' phase (see Algorithm \ref{Algo_LOExplore}) aims to find new local optimal solutions of increasing quality. This is achieved by a combined strategy mixing a variable neighborhood descent (VND) procedure (line 6, Alg. \ref{Algo_LOExplore}, see Section \ref{Descent}) and a tabu search (TS) procedure (line 7, Alg. \ref{Algo_LOExplore}, see Section \ref{Tabu}). For each VND-TS run (each `while' iteration), the VND procedure exploits, with the best-improvement strategy, two neighborhoods $N_1$ and $N_2$ to locate a local optimal solution. Then from this solution, the TS procedure is triggered to examine additional local optimal solutions with another neighborhood $N_3$. At the end of TS, its best solution ($S_c$) is used to update the recorded best solution ($S_b$) found during the current VND-TS run, while its last solution ($S$) is used as the new starting point of the next iteration of the `Explore' phase. The `Explore' phase terminates when the best solution ($S_b$) found during this run cannot be updated during $\lambda_{max}$ consecutive iterations ($\lambda_{max}$ is a parameter called \textit{exploration depth}).

\subsubsection{Variable Neighborhood Descent Search}
\label{Descent}

\begin{algorithm}
\footnotesize
\caption{Variable Neighborhood Descent - VND}\label{Algo_Descent}
\begin{algorithmic}[1]
   \STATE \textbf{Input}: Input solution $S$, neighborhoods $N_1$ and $N_2$, sampling probability $\rho$.
   \STATE \textbf{Output}: The best solution $S_b$ found during the VND search.
   \STATE $S_b \gets S$     \hfill	/*$S_b$ record the best solution found so far*/
   \STATE $Improve \gets True$
   \WHILE{$Improve$}
	 \STATE $S \gets argmax\{f(S'): S' \in N_1(S)\}$
   \IF{$f(S) > f(S_b)$}
   \STATE $S_b \gets S$   \hfill	/*Update the best solution found so far*/
   \STATE $Improve = True$
   \ELSE
   \STATE $N_2^- \gets Sampling(N_2,S,\rho)$
	 \STATE $S \gets argmax\{f(S'): S' \in N_2^-(S) \}$ 
	 \IF {$f(S) > f(S_b)$}
   \STATE $S_b \gets S$   \hfill	/*Update the best solution found so far*/
   \STATE $Improve \gets True$
   \ELSE
   \STATE $Improve = False$
   \ENDIF
   \ENDIF
   \ENDWHILE  
   \RETURN $S_b$
\end{algorithmic}
\end{algorithm}

\begin{algorithm}
\footnotesize
\caption{Sampling Procedure}\label{Algo_Sampling}
\begin{algorithmic}[1]
   \STATE \textbf{Input}: Input solution $S$, neighborhood $N_2$, sampling probability $\rho$.
   \STATE \textbf{Output}: Set $N_2^-$ of sampled solutions from $N_2(S)$
   \STATE  $N_2^- \gets \emptyset$   
	 \FOR {each $S' \in$ $N_2(S)$}
   \IF { $random() < \rho$ }
	 \STATE  $N_2^- \gets N_2^- \cup \{S'\}$
   \ENDIF
   \ENDFOR
   \RETURN $N_2^-$
\end{algorithmic}
\end{algorithm}

Following the general variable neighborhood descent search \cite{MladenovccHansen1997}, the VND procedure (Algorithm \ref{Algo_Descent})  relies on two neighborhoods ($N_1$ and $N_2$, see Sections \ref{Operator}) to explore the search space. Specifically, VND examines the neighborhood $N_1$ at first and iteratively identifies a best-improving neighbor solution in $N_1$ to replace the current solution. When a local optimal solution is reached within $N_1$, VND switches to the neighborhood $N_2$. As we explain in Section \ref{Operator}, given the large size of $N_2$, VND only examines a subset $N_2^-$ which is composed of $\rho \times |N_2|$ randomly solutions of $N_2$ ($\rho$ is a parameter called sampling probability and Algorithm \ref{Algo_Sampling} shows the sampling procedure where $random()$ is a random real number in [0,1]). If an improving neighbor solution is detected in $N_2^-$, VND switches back to $N_1$. VND terminates when no improving solution can be found within both neighborhoods. In Section \ref{Strategy1}, we study the influence of this sampling strategy.

\subsubsection{Move Operators, Neighborhoods and VND Exploration}
\label{Operator}

To explore candidate solutions of the search space, the I2PLS algorithm employs the general $swap$ operator to transform solutions. Specifically, let $S=<A,\bar{A}>$ be a given solution with $A$ and $\bar{A}$ being the set of selected and non-selected items. Let $swap(q,p)$ denote the operation that deletes $q$ items from $A$ and adds $p$ other items from $\bar{A}$ into $A$. By limiting $q$ and $p$ to specific values, we introduce two particular $swap(q,p)$ operators.

The first operator $swap_1(q,p) \ (q \in \{0,1\}, \ p = 1)$ includes two customary operations as described in the literature \cite{lai2018two,WuHao2015,zhouHaoGoeffon2017}: the $Add$ operator and the $Exchange$ operator. Basically, $swap_1(q,p)$ either adds an item from $\bar{A}$ into $A$ or exchanges one item in $A$ with another item in $\bar{A}$ while keeping the capacity constraint satisfied. 

The second operator $swap_2(q,p) \ (3 \leq q + p \leq 4)$ covers three different cases: delete two items from $A$ and add one item from $\bar{A}$ into $A$; delete one item from $A$ and add two items from $\bar{A}$ into $A$; exchanges two items of $A$ against two items of $\bar{A}$. These three operations are subject to the capacity constraint.

On the basis of these two swap operators, we define the neighborhoods $N_1^w$ and $N_2^w$ induced by $swap_1$ and $swap_2$ as follows.

\begin{equation}\label{N_1}
N_1^w(S) = \{S': S'=S \oplus swap_1(q,p), q \in \{0,1\}, \ p = 1, \sum\limits_{j \in \cup_{i \in S'} U_i} w_j \leq C\}
\end{equation}	
	
\begin{equation}\label{N_2}
N_2^w(S) = \{S': S'=S \oplus swap_2(q,p), 3 \leq q + p \leq 4, \sum\limits_{j \in \cup_{i \in S'} U_i} w_j \leq C\}
\end{equation}	
	
where $S'=S \oplus swap_k(q,p)$ ($k=1,2$) is the neighbor solution of the incumbent solution $S$ obtained by applying $swap_1(q,p)$ or $swap_2(q,p)$ to $S$.

$N_1^w$ and $N_2^w$ are bounded in size by $O(|A| \times |\bar{A}|)$ and $O(\binom 2 {|A|} \times \binom 2 {|\bar{A}|})$ respectively. 

Given the large sizes of these neighborhoods, it is obvious that exploring all the neighbor solutions at each iteration will be very time consuming. To cope with this problem, we adopt the idea of a filtering strategy that excludes the non-promising neighbor solutions from consideration \cite{lai2018two}. Specifically, a neighbor solution $S'$ qualifies as promising if $f(S') > f(S_b)$ holds, where $S_b$ is the best solution found so far in Algorithm \ref{Algo_Descent}. Using this filtering strategy, we define the following reduced neighborhoods $N_1$ and $N_2$.

\begin{equation}\label{N_1}
N_1(S) = \{S'\in N_1^w(S): f(S') > f(S_b)\}
\end{equation}	

\begin{equation}\label{N_1}
N_2(S) = \{S'\in N_2^w(S): f(S') > f(S_b)\}
\end{equation}	

As explained in Section \ref{Descent} and Algorithm \ref{Algo_Descent}, the VND procedure successively examines solutions of these two neighborhoods $N_1$ and $N_2$. Notice that $swap_2$ leads generally to a very large number of neighbor solutions such that even the reduced neighborhood $N_2$ can still be too large to be explored efficiently. For this reason, the VND procedure explores a sampled portion of $N_2$ at each iteration, according to the sampling procedure shown in Algorithm \ref{Algo_Sampling}.

\subsubsection{Tabu Search}
\label{Tabu}

\begin{algorithm}
\footnotesize
\caption{Tabu Search - TS}\label{Algo_Tabu}
\begin{algorithmic}[1]
   \STATE \textbf{Input}: Input solution $S$, Neighborhood $N_3$, tabu search depth $\omega_{max}$
   \STATE \textbf{Output}: The best solution $S_b$ found during tabu search, the last solution $S$ of tabu search.
   \STATE $S_b \gets S$     \hfill	/*$S_b$ records the best solution found so far*/
	\STATE $\omega \gets 0$   \hfill	/*$\omega$ counts the number of consecutive non-improving iterations */  
    \WHILE {$\omega < \omega_{max}$}
	 \STATE $S \gets argmax\{f(S'): S' \in N_3(S)\ $and$\ S' $\ is\ not\ forbidden\ by\ the$\ tabu\_list \}$
   \IF    {$f(S) > f(S_b)$}
   \STATE $S_b \gets S$ \hfill /* Update the best solution $S_b$ found so far */ 
   \STATE $\omega \gets 0$
   \ELSE  
   \STATE $\omega \gets \omega + 1$
   \ENDIF
   	 \STATE Update the $tabu\_list$
   \ENDWHILE
   \RETURN $(S_b,S)$
\end{algorithmic}
\end{algorithm}

To discover still better solutions when the VND search terminates, we trigger the tabu search (TS) procedure (Algorithm \ref{Algo_Tabu}) that is adapted from the general tabu search metaheuristic \cite{glover1997tabu}. To explore candidate solutions, TS relies on the $swap_3(q,p) \ ( 1 \leq p + q \leq 2)$ operator, which extends $swap_1$ used in VND by including the case $q = 1, p = 0$, which corresponds to the drop operation (i.e., deleting an item from $A$ without adding any new item). One notices that $swap_3(1,0)$ always leads to a neighbor solution of worse quality, which can be usefully selected for search diversification. We use $N_3$ to denote the neighborhood induced by $swap_3$.

\begin{equation}\label{N_1}
N_3(S) = \{S': S'=S \oplus swap_3(q,p), 1 \leq p + q \leq 2, \sum\limits_{j \in \cup_{i \in S'} U_i} w_j \leq C\}
\end{equation}	

As shown Algorithm \ref{Algo_Tabu}, the TS procedure iteratively makes transitions from the incumbent solution $S$ to a selected neighbor solution $S'$ in $N_3$. At each iteration, TS selects the best neighbor solution $S'$ in $N_3$ (or one of the best ones if there are multiple best solutions) that is not forbidden by the so-called tabu list ($tabu\_list$) (line 6, Alg. \ref{Algo_Tabu}, see below). Notice that if no improving solution exists in $N_3(S)$, the selected neighbor solution $S'$ is necessarily a worsening or equal-quality solution relative to $S$. It is this feature that allows TS to go beyond local optimal traps. To prevent the search from revisiting previously encountered solutions, the tabu list is used to record the items involved in the swap operation. And each item $i$ of the tabu list is then forbidden to take part in any swap operation during the next $T_i$ consecutive iterations where $T_i$ is called the tabu tenure of item $i$ and is empirically fixed as follows. 

\begin{equation}\label{T_i}
\quad \quad \quad T_i = \begin{cases}
		0.4 \times |A|, & $if$ \ i \in A; \\
		0.2 \times |\bar{A}| \times (100 / m) , &  $if$\ i \in \bar{A}.
		\end{cases}
\end{equation}

TS terminates when its best solution cannot be further improved during $\omega_{max}$ consecutive iterations ($\omega_{max}$ is a parameter called the tabu search depth).

\subsection{Frequency-Based Local Optima Escaping Phase}
\label{Escaping}

The `Explore' phase aims to diversify the search by exploring new search regions. For this purpose, the algorithm keeps track of the frequencies that each item has been displaced and uses the frequency information to modify (perturb) the incumbent solution. Particularly, we adopt an integer vector $F$ of length $m$ whose elements are initialized to zero. Each time an item $i$ is displaced by a swap operation, $F_i$ is increased by one. Thus, items with a low frequency are those that are not frequently moved during the 'Explore' phase. Then when the `Explore' phase terminates and before the next round of the `Explore' phase starts, we modify the best  solution $S_b=<A_b,\bar{A_b}>$ as follows. We delete the top $\eta \times |A_b|$ least frequently moved items from $A_b$ ($\eta $ is a parameter called \textit{perturbation strength} and adds to $A_b$ randomly select items from $\bar{A_b}$ until the knapsack capacity is reached. This perturbed solution serves as the new starting solution $S_0$ of the next iteration of the algorithm (see line 10, Alg. \ref{Algo_I2PLS}). In Section \ref{Strategy2}, we study the usefulness of this perturbation strategy. 

\section{Experimental Results and Comparisons}
\label{Sec_Results}

This section presents a performance assessment of the I2PLS algorithm. We show computational results on the 30 benchmark instances commonly used in the literature, in comparison with three state-of-the-art algorithms for SUKP. We also present the first results from the CPLEX solver. 

\subsection{Benchmark Instances}
\label{Instance}

We use the 30 benchmark instances provided in \cite{he2018novel}, which were also tested in 2 other recent studies \cite{ozsoydan2018swarm,fengetal2019}. These instances are divided into three sets according to the relationship between the number of items and elements ranging from 85 to 500, where each instance has a different density $\alpha$ of elements in an item and a different ratio $\beta$ of the knapsack capacity to the total weight of all elements. Let $R$ be a $m \times n$ binary relation matrix between $m$ items and $n$ elements where $R_{ij} = 1$ indicates the presence of element $j$ in item $i$, $w_j$ be the weight of element $j$, and $C$ the knapsack capacity. Then $m\_n\_\alpha\_\beta$ designates an instance with $m$ items and $n$ elements, density of $\alpha$ and ratio of $\beta$, where $\alpha = (\sum_{i=1}^m \sum_{j=1}^n R_{ij}) / (mn)$ and $\beta = C / \sum_{j=1}^n w_j$. The characteristics of the three sets of instances are shown in Tables \ref{Results1} to \ref{Results3}. 

\subsection{Experimental Setting and reference algorithms}
\label{Setting}

The proposed algorithm was implemented in C++ and compiled using the g++ compiler with the -O3 option. The experiments were carried on an Intel Xeon E5-2670 processor with 2.5 GHz and 2 GB RAM under the Linux operating system. 

\renewcommand{\baselinestretch}{0.9}\large\normalsize
\begin{table}[h]\centering
\begin{scriptsize}
\caption{Settings of parameters.}
\label{Parameter_Settings}
\begin{tabular}{cclc}
\hline
Parameters & Sect. & Description & Value\\
\hline
$\lambda_{max}$ &    \ref{Algo_LOExplore}                        &  Exploration depth        & 2 \\
$\rho$     & \ref{Descent}                        &  Sampling probability for VND & 5 \\
$\omega_{max}$     & \ref{Tabu}                        & Tabu search depth & 100 \\
$\eta$     & \ref{Escaping}                        & Perturbation strength in escaping phase &  0.5  \\
\hline
\end{tabular}
\end{scriptsize}
\end{table}
\renewcommand{\baselinestretch}{1.0}\large\normalsize

Table \ref{Parameter_Settings} shows the setting of parameters used in our algorithm, whose values were discussed in Section \ref{Analysis of Parameters}. Given the stochastic nature of the algorithm, we ran 100 times (like in \cite{he2018novel,ozsoydan2018swarm}) with different random seeds to solve each instance, with a cut-off time of 500 seconds per run.

For the comparative studies, we use as reference algorithms the following three very recent algorithms: BABC (binary artificial bee colony algorithm) (2018), which is the best performing among five population-based algorithms tested in \cite{he2018novel} (2018), gPSO (binary particle swarm optimization algorithm) (2019) \cite{ozsoydan2018swarm} and MA (discrete moth search algorithm) \cite{fengetal2019}. Among these reference algorithm, we obtained the code of BABC. So for BABC, we report both the results listed in \cite{he2018novel} as well as the results by running the BABC code on our computer under the same time limit of 500 seconds. For gPSO and MA, we cite the results reported in the corresponding papers. The results of these reference algorithms have been obtained on computing platforms with the following features: an Intel Core i5-3337u processor with 1.8 GHz and 4 GB RAM for BABC, an Intel Core i7-4790K 4.0 GHz processor with 32 GB RAM for gPSO, and an Intel Core i7-7500 processor with 2.90 GHz and 8.00 GB RAM for MA. 

Additionally, we notice that until now, no result has been reported by using the general integer linear programming (ILP) approach to solve the SUKP. Therefore, we include in our experimental study the results achieved by the ILP CPLEX solver (version 12.8) under a time limit of 2 hours based on the 0/1 linear programming model presented in the Appendix.

\subsection{Computational Results and Comparisons}
\label{Result}

The computational results of I2PLS on the three sets of benchmark instances are reported in Tables 2-4, together with the results of the reference algorithms (BABC \cite{he2018novel}, gPSO \cite{ozsoydan2018swarm}, MS \cite{fengetal2019}) where BABC* corresponds to the results by running the BABC code as explained in Section \ref{Setting}. The first column of each table gives the name of each instance. Column 2 (Best\_Known) indicates the best known value reported in the literature and compiled from \cite{fengetal2019,he2018novel,ozsoydan2018swarm}. The best lower bound (LB) and upper bound (UB) achieved by the CPLEX solver are given in columns 3 and 4. Column 5 lists respectively the four performance indicators: best objective value ($f_{best}$), average objective value over 100 runs ($f_{avg}$), standard deviations over 100 runs ($std$), and average run times $t_{avg}$ in seconds to reach the best objective value. Columns 6 to 9 present the computational statistics of the compared algorithms. The best values of $f_{best}$ and $f_{avg}$ among the results of the compared algorithms are highlighted in bold and the equal values are indicated in italic. Entries with "-" mean that the results are not available.

Given the fact that the compared algorithms were run on different computing platforms and they report solutions of various quality, it is not meaningful to compare the computation times. Therefore, the comparisons are mainly based on the quality, while run times (when they are available) are included only for indicative purposes. 

Finally, Table \ref{Summary_result} provides a summary of all the algorithms on all 30 benchmark instances where rows $\# Better$, $\# Equal$ and $\# Worse$ indicate the number of instances for which each algorithm obtains a better, equal or worse $f_{best}$ value compared to the best-known values in the literature (Best$\_$Known). Moreover, to further analyze the performance of our I2PLS algorithm, we use the non-parametric Wilcoxon signed-rank test to check the statistical significance of the compared results between I2PLS and each reference algorithm in terms of $f_{best}$ values. The outcomes of the Wilcoxon tests are shown in the last row of Table \ref{Summary_result} where a $p$-$value$ smaller than 0.05 implies a significant performance difference between I2PLS and its competitor.

\begin{table}[!htp]\centering
	\caption{Computational results and comparison of the proposed I2PLS algorithm with the reference algorithms on the first set of instances ($m>n$).} 
	\renewcommand{\baselinestretch}{0.9}\large\normalsize
	\begin{scriptsize}
	%\centering
	\setlength{\tabcolsep}{1mm}{
	\begin{tabular}{lccc|cccccc}
	\toprule[0.75pt]

Instance & \multicolumn{1}{c}{Best$\_$Known} & \multicolumn{1}{c}{LB}  & \multicolumn{1}{c}{UB} &  Results & BABC & BABC* & gPSO & MSO4 & I2PLS \\

\hline

100\_85\_0.10\_0.75$^*$ & \textit{13283} & \textit{13283} & \textit{13283} & \textit{$f_{best}$} & 13251 & \textit{13283} & \textit{13283} & \textit{13283} & \textit{13283} \\
&  &  &  & \textit{$f_{avg}$} & 13208.5 & \textit{13283} & 13050.53 & 13062 & \textit{13283} \\
&  &  &  & \textit{$std$}  & 92.63 & 0 & 37.41 & - & 0 \\
&  &  &  & \textit{$t_{avg}$} & 0.210 & 51.102 & - & - & 3.094 \\
\hline
100\_85\_0.15\_0.85$^*$ & 12274 & \textit{12479} & \textit{12479} & \textit{$f_{best}$} & 12238 & \textit{12479} & 12274 & - & \textit{12479} \\
&  &  &  & \textit{$f_{avg}$} & 12155 & \textbf{12479} & 12084.82 & - & 12335.13 \\
&  &  &  & \textit{$std$}  & 53.29 & 0 & 95.38 & - & 98.78 \\
&  &  &  & \textit{$t_{avg}$} & 0.223 & 24.032 & - & - & 103.757 \\
\hline
200\_185\_0.10\_0.75 & \textit{13521} & 11585 & 27055.82 & \textit{$f_{best}$} & 13241 & 13402 & 13405 & \textit{13521} & \textit{13521} \\
&  &  &  & \textit{$f_{avg}$} & 13064.4 & 13260.16 & 13286.56 & 13193 & \textbf{13521} \\
&  &  &  & \textit{$std$}  & 99.57 & 38.98 & 93.18 & - & 0 \\
&  &  &  & \textit{$t_{avg}$} & 1.562 & 253.693 & - & - & 71.984 \\
\hline
200\_185\_0.15\_0.85 &  14044 & 11017 & 29625.82 & \textit{$f_{best}$} & 13829 & \textit{14215} & 14044 & - & \textit{14215} \\
&  &  &  & \textit{$f_{avg}$} & 13359.2 & 14026.18 & 13492.60 & - & \textbf{14031.28} \\
&  &  &  & \textit{$std$}  & 234.99 & 151.55 & 328.72 & - & 131.46 \\
&  &  &  & \textit{$t_{avg}$} & 1.729 & 241.932 & - & - & 180.809 \\
\hline
300\_285\_0.10\_0.75 &  11335 & 9028 & 43937.51 & \textit{$f_{best}$} & 10428 & 10572 & 11335 & 11127 & \textbf{11563} \\
&  &  &  & \textit{$f_{avg}$} & 9994.76 & 10466.45 & 10669.51 & 10302 & \textbf{11562.02} \\
&  &  &  & \textit{$std$}  & 154.03 & 61.94 & 227.85 & - & 3.94 \\
&  &  &  & \textit{$t_{avg}$} & 5.281 & 315.240 & - & - & 181.248 \\
\hline
300\_285\_0.15\_0.85 &  12245 & 6889 & 53164.23 & \textit{$f_{best}$} & 12012 & 12245 & 12245 & - & \textbf{12607} \\
&  &  &  & \textit{$f_{avg}$} & 10902.9 & 12019.28 & 11607.10 & - & \textbf{12364.55} \\
&  &  &  & \textit{$std$}  & 449.45 & 85.76 & 477.80 & - & 83.03 \\
&  &  &  & \textit{$t_{avg}$} & 5.673 & 226.818 & - & - & 240.333 \\
\hline
400\_385\_0.10\_0.75 &  \textit{11484} & 8993 & 66798.30 & \textit{$f_{best}$} & 10766 & 11021 & \textit{11484} & 11435 & \textit{11484} \\
&  &  &  & \textit{$f_{avg}$} & 10065.2 & 10608.91 & 10915.87 & 10411 & \textbf{11484} \\
&  &  &  & \textit{$std$}  & 241.45 & 138.07 & 367.75 & - & 0 \\
&  &  &  & \textit{$t_{avg}$} & 12.976 & 293.560 & - & & 31.801 \\
\hline
400\_385\_0.15\_0.85 &  10710 & 5179 & 77480.39 & \textit{$f_{best}$} & 9649 & 9649 & 10710 & - & \textbf{11209} \\
&  &  &  & \textit{$f_{avg}$} & 9135.98 & 9503.65 & 9864.55 & - & \textbf{11157.26} \\
&  &  &  & \textit{$std$}  & 151.90 & 94.69 & 315.38 & - & 87.29 \\
&  &  &  & \textit{$t_{avg}$} & 13.359 & 270.813 & - & - & 141.525 \\
\hline
500\_485\_0.10\_0.75 &  11722 & 7202 & 86166.50 & \textit{$f_{best}$} & 10784 & 10927 & 11722 & 11031 & \textbf{11771} \\
&  &  &  & \textit{$f_{avg}$} & 10452.2 & 10628.31 & 11184.51 & 10716 & \textbf{11729.76} \\
&  &  &  & \textit{$std$}  & 114.35 & 70.3135 & 322.98 & - & 6.59 \\
&  &  &  & \textit{$t_{avg}$} & 25.372 & 486.210 & - & - & 349.545 \\
\hline
500\_485\_0.15\_0.85 & 10022 & 4762 & 97218.01 & \textit{$f_{best}$} & 9090 & 9306 & 10022 & - & \textbf{10238} \\
&  &  &  & \textit{$f_{avg}$} & 8857.89 & 9014.01 & 9299.56 & - & \textbf{10133.94} \\
&  &  &  & \textit{$std$}  & 94.55 & 64.06 & 277.62 & - & 94.72 \\
&  &  &  & \textit{$t_{avg}$} & 26.874 & 482.740 & - & - & 369.375 \\

	\bottomrule[0.75pt]
	\end{tabular}
	}
\end{scriptsize}
\label{Results1}
\end{table}
\renewcommand{\baselinestretch}{1.0}\large\normalsize

\begin{table}[!htp]\centering
	\caption{Computational results and comparison of the proposed I2PLS algorithm with the reference algorithms on the second set of instances ($m=n$).} 
	\renewcommand{\baselinestretch}{0.9}\large\normalsize
\begin{scriptsize}
	\setlength{\tabcolsep}{1mm}{
	\begin{tabular}{lccc|cccccc}
	\toprule[0.75pt]

Instance & \multicolumn{1}{c}{Best$\_$Known} & \multicolumn{1}{c}{LB}  & \multicolumn{1}{c}{UB} & Results & BABC & BABC* & gPSO & MSO4 & I2PLS \\
\hline

100\_100\_0.10\_0.75$^*$ & \textit{14044} & \textit{14044} &  \textit{14044} & \textit{$f_{best}$} & 13860 & \textit{14044} & \textit{14044} & \textit{14044} & \textit{14044} \\
&  &  &  & \textit{$f_{avg}$} & 13734.9 & 14040.87 & 13854.71 & 13649 & \textbf{14044} \\
&  &  &  & \textit{$std$}  & 70.76 & 11.51 & 96.23 & - & 0 \\
&  &  &  & \textit{$t_{avg}$} & 0.213 & 169.848 & - & - & 38.245 \\
\hline
100\_100\_0.15\_0.85$^*$ & \textit{13508} & \textit{13508} & \textit{13508} & \textit{$f_{best}$} & \textit{13508} & \textit{13508} & \textit{13508} & - & \textit{13508} \\
&  &  &  & \textit{$f_{avg}$} & 13352.4 & \textbf{13508} & 13347.58 & - & 13451.50 \\
&  &  &  & \textit{$std$}  & 155.14 & 0 & 194.34 & - & 126.49 \\
&  &  &  & \textit{$t_{avg}$} & 0.244 & 6.795 & - & - & 70.587 \\
\hline
200\_200\_0.10\_0.75 &  \textit{12522} & 11187 & 29394.32 & \textit{$f_{best}$} & 11846 & 12350 & \textit{12522} & 12350 & \textit{12522} \\
&  &  &  & \textit{$f_{avg}$} & 11194.3 & 11953.11 & 11898.73 & 11508 & \textbf{12522} \\
&  &  &  & \textit{$std$}  & 249.58 & 97.57 & 391.83 & - & 0 \\
&  &  &  & \textit{$t_{avg}$} & 1.633 & 183.130 & - & - & 54.780 \\
\hline
200\_200\_0.15\_0.85 &  \textit{12317} & 9258 & 30610.99 & \textit{$f_{best}$} & 11521 & 11929 & \textit{12317} & - & \textit{12317} \\
&  &  &  & \textit{$f_{avg}$} & 10945 & 11695.21 & 11584.64 & - & \textbf{12280.07} \\
&  &  &  & \textit{$std$}  & 255.14 & 78.33 & 275.32 & - & 57.77 \\
&  &  &  & \textit{$t_{avg}$} & 1.819 & 147.930 & - & - & 238.348 \\
\hline
300\_300\_0.10\_0.75 &  12736 & 11007 & 45191.75 & \textit{$f_{best}$} & 12186 & 12304 & 12695 & 12598 & \textbf{12817} \\
&  &  &  & \textit{$f_{avg}$} & 11945.8 & 12202.80 & 12411.27 & 11541 & \textbf{12817} \\
&  &  &  & \textit{$std$}  & 127.80 & 67.81 & 225.80 & - & 0 \\
&  &  &  & \textit{$t_{avg}$} & 5.315 & 202.515 & - & - & 66.403 \\
\hline
300\_300\_0.15\_0.85 &  11425 & 7590 & 51891.53 & \textit{$f_{best}$} & 10382 & 10857 & 11425 & - & \textbf{11585} \\
&  &  &  & \textit{$f_{avg}$} & 9859.69 & 10383.64 & 10568.41 & - & \textbf{11512.18} \\
&  &  &  & \textit{$std$}  & 177.02 & 75.79 & 327.48 & - & 73.15 \\
&  &  &  & \textit{$t_{avg}$} & 6.019 & 113.380 & - & - & 220.100 \\
\hline
400\_400\_0.10\_0.75 &  11531 & 7910 & 68137.98 & \textit{$f_{best}$} & 10626 & 10869 & 11531 & 10727 & \textbf{11665} \\
&  &  &  & \textit{$f_{avg}$} & 10101.1 & 10591.65 & 10958.96 & 10343 & \textbf{11665} \\
&  &  &  & \textit{$std$}  & 196.99 & 105.83 & 274.90 & - & 0 \\
&  &  &  & \textit{$t_{avg}$} & 12.805 & 298.970 & - & - & 18.733 \\
\hline
400\_400\_0.15\_0.85 &  10927 & 4964 & 77719.78 & \textit{$f_{best}$} & 9541 & 10048 & 10927 & - & \textbf{11325} \\
&  &  &  & \textit{$f_{avg}$} & 9032.95 & 9602.13 & 9845.17 & - & \textbf{11325} \\
&  &  &  & \textit{$std$}  & 194.18 & 142.77 & 358.91 & - & 0 \\
&  &  &  & \textit{$t_{avg}$} & 12.953 & 386.555 & - & - & 76.000 \\
\hline
500\_500\_0.10\_0.75 &  10888 & 7500 & 85184.48 & \textit{$f_{best}$} & 10755 & 10755 & 10888 & 10355 & \textbf{11249} \\
&  &  &  & \textit{$f_{avg}$} & 10328.5 & 10522.56 & 10681.46 & 9919 & \textbf{11243.40} \\
&  &  &  & \textit{$std$}  & 94.615 & 70.17 & 125.36 & - & 27.43 \\
&  &  &  & \textit{$t_{avg}$} & 27.735 & 194.490 & - & - & 134.186 \\
\hline
500\_500\_0.15\_0.85 &  10194 & 3948 & 101964.36 & \textit{$f_{best}$} & 9318 & 9601 & 10194 & - & \textbf{10381} \\
&  &  &  & \textit{$f_{avg}$} & 9180.74 & 9334.52 & 9703.62 & - & \textbf{10293.89} \\
&  &  &  & \textit{$std$}  & 84.91 & 40.59 & 252.84 & - & 85.53 \\
&  &  &  & \textit{$t_{avg}$} & 27.813 & 135.130 & - & - & 237.894 \\

	\bottomrule[0.75pt]
	\end{tabular}}

\label{Results2}
\end{scriptsize}
\end{table} 
\renewcommand{\baselinestretch}{1.0}\large\normalsize

\begin{table}[!htp]\centering
	\caption{Computational results and comparison of the proposed I2PLS algorithm with the reference algorithms on the third set of instances ($m<n$).} 
	\renewcommand{\baselinestretch}{0.9}\large\normalsize
	\begin{scriptsize}	
	\setlength{\tabcolsep}{1mm}{
	\begin{tabular}{lccc|cccccc}
	\toprule[0.75pt]

Instance & \multicolumn{1}{c}{Best$\_$Known}  & \multicolumn{1}{c}{LB}  & \multicolumn{1}{c}{UB} & Results & BABC & BABC* & gPSO & MSO4 & I2PLS \\
\hline

85\_100\_0.10\_0.75$^*$  & \textit{12045} & \textit{12045} & \textit{12045} & \textit{$f_{best}$} & 11664 & \textit{12045} & \textit{12045} & 11735 & \textit{12045} \\
&  &  &  & \textit{$f_{avg}$} & 11182.7 & 11995.12 & 11486.95 & 11287 & \textbf{12045} \\
&  &  &  & \textit{$std$}  & 183.57 & 53.15 & 137.52 & - & 0 \\
&  &  &  & \textit{$t_{avg}$} & 0.188 & 206.570 & - & - & 2.798 \\
\hline
85\_100\_0.15\_0.85$^*$ & \textit{12369} & \textit{12369} & \textit{12369} & \textit{$f_{best}$} & \textit{12369} & \textit{12369} & \textit{12369} & - & \textit{12369} \\
&  &  &  & \textit{$f_{avg}$} & 12081.6 & \textbf{12369} & 11994.36 & - & 12315.53 \\
&  &  &  & \textit{$std$}  & 193.79 & 0 & 436.81 & - & 62.60 \\
&  &  &  & \textit{$t_{avg}$} & 0.217 & 0.5313 & - & - & 17.47 \\
\hline
185\_200\_0.10\_0.75 & \textit{13696} & 12264 & 25702.48 & \textit{$f_{best}$} & 13047 & 13647 & \textit{13696} & 13647 & \textit{13696} \\
&  &  &  & \textit{$f_{avg}$} & 12522.8 & 13179.14 & 13204.26 & 13000 & \textbf{13695.60} \\
&  &  &  & \textit{$std$}  & 201.35 & 100.78 & 366.56 & - & 3.68 \\
&  &  &  & \textit{$t_{avg}$} & 1.502 & 202.560 & - & - & 124.136 \\
\hline
185\_200\_0.15\_0.85 & 11298 & 8608 & 26289.16 & \textit{$f_{best}$} & 10602 & 10926 & \textit{11298} & - & \textit{11298} \\
&  &  &  & \textit{$f_{avg}$} & 10150.6 & 10749.46 & 10801.41 & - & \textbf{11276.17} \\
&  &  &  & \textit{$std$}  & 152.91 & 97.24 & 205.76 & - & 83.78 \\
&  &  &  & \textit{$t_{avg}$} & 1.948 & 259.050 & - & - & 139.865 \\
\hline
285\_300\_0.10\_0.75 & 11568 & 9421 & 44274.85 & \textit{$f_{best}$} & 11158 & 11374 & \textit{11568} & 11391 & \textit{11568} \\
&  &  &  & \textit{$f_{avg}$} & 10775.9 & 11143.69 & 11317.99 & 10816 & \textbf{11568} \\
&  &  &  & \textit{$std$}  & 116.80 & 76.90 & 182.82 & - & 0 \\
&  &  &  & \textit{$t_{avg}$} & 5.450 & 426.680 & - & - & 25.128 \\
\hline
285\_300\_0.15\_0.85 & 11517 & 7634 & 51440.30 & \textit{$f_{best}$} & 10528 & 10822 & 11517 & - & \textbf{11802} \\
&  &  &  & \textit{$f_{avg}$} & 9897.92 & 10396.60 & 10899.20 & - & \textbf{11790.43} \\
&  &  &  & \textit{$std$}  & 186.53 & 128.6345 & 300.36 & - & 27.51 \\
&  &  &  & \textit{$t_{avg}$} & 5.571 & 192.575 & - & - & 206.422 \\
\hline
385\_400\_0.10\_0.75 & 10483 & 9591 & 59917.77 & \textit{$f_{best}$} & 10085 & 10110 & 10483 & 9739 & \textbf{10600} \\
&  &  &  & \textit{$f_{avg}$} & 9537.5 & 9926.18 & 10013.43 & 9240 & \textbf{10536.53} \\
&  &  &  & \textit{$std$}  & 184.62 & 87.43 & 202.40 & - & 56.08 \\
&  &  &  & \textit{$t_{avg}$} & 13.012 & 203.870 & - & - & 234.475 \\
\hline
385\_400\_0.15\_0.85 & 10338 & 5810 & 73409.01 & \textit{$f_{best}$} & 9456 & 9659 & 10338 & - & \textbf{10506} \\
&  &  &  & \textit{$f_{avg}$} & 9090.03 & 9444.34 & 9524.98 & - & \textbf{10502.64} \\
&  &  &  & \textit{$std$}  & 156.69 & 46.40 & 286.16 & - & 23.52 \\
&  &  &  & \textit{$t_{avg}$} & 13.724 & 177.910 & - & - & 129.505 \\
\hline
485\_500\_0.10\_0.75 & 11094 & 5940 & 84239.56 & \textit{$f_{best}$} & 10823 & 10835 & 11094 & 10539 & \textbf{11321} \\
&  &  &  & \textit{$f_{avg}$} & 10483.4 & 10789.57 & 10687.62 & 10190 & \textbf{11306.47} \\
&  &  &  & \textit{$std$}  & 228.34 & 27.29 & 168.06 & - & 36.00 \\
&  &  &  & \textit{$t_{avg}$} & 27.227 & 299.260 & - & - & 207.118 \\
\hline
485\_500\_0.15\_0.85 & 10104 & 4325 & 100374.77 & \textit{$f_{best}$} & 9333 & 9380 & 10104 & - & \textbf{10220} \\
&  &  &  & \textit{$f_{avg}$} & 9085.57 & 9258.82 & 9383.28 & - & \textbf{10179.45} \\
&  &  &  & \textit{$std$}  & 115.62 & 58.72 & 241.01 & - & 46.97 \\
&  &  &  & \textit{$t_{avg}$} & 28.493 & 49.170 & - & - & 238.630 \\

	\bottomrule[0.75pt]
	\end{tabular}}

\label{Results3}
\end{scriptsize}
\end{table}
\renewcommand{\baselinestretch}{1.0}\large\normalsize

\begin{table}[!htp]\centering
	\caption{Summary of numbers of instances for which each algorithm reports a better, equal or worse $f_{bst}$ value compared to the best-known value in the literature and $p$-$values$ of the Wilcoxon singned-rank test on $f_{best}$ values over all instances between I2PLS and each reference algorithm including the best-known values.} 
	\renewcommand{\baselinestretch}{0.9}\large\normalsize
	\begin{scriptsize}
	\setlength{\tabcolsep}{1mm}{
	\begin{tabular}{lcccccc}
	\toprule[0.75pt]

Instance  & Best$\_$Known & BABC & BABC* & gPSO & MSO4 & I2PLS \\
\hline

\# Better & - & 0 & 2 & 0 & 0 & 18 \\
 
\# Equal & - & 2 & 6 & 28 & 3 & 12 \\

\# Worse & - & 28 & 22 & 2  & 12 & 0 \\

$p$-$value$ & 2.14e-4 & 4.00e-6 & 2.89e-5 & 1.43e-4 & 2.52e-3 & - \\

	\bottomrule[0.75pt]
	\end{tabular}}

\label{Summary_result}
\end{scriptsize}
\end{table} 
\renewcommand{\baselinestretch}{1.0}\large\normalsize

From Tables \ref{Results1} to \ref{Results3}, we observe that our I2PLS algorithm performs extremely well compared to the state-of-the-art results on the set of 30 benchmark instances. In particular, I2PLS improves on the best-known results of the literature for 18 out of 30 instances, while matching the best known-results for the remaining 12 instances. Notice that among these 11 instances, 6 instances with 85 and 100 items are solved to optimality by CPLEX (LB=UB), which are indeed not challenging for the other algorithms. Compared to the reference algorithms (BABC/BABC*, gPSO, MS) I2PLS reports better or equal $f_{best}$ values for all the tested instances without exception. In terms of the average results ($f_{avg}$), I2PLS also performs very well by reporting better or equal $f_{best}$ values for all instances except three cases ($100\_85\_0.15\_0.85$, $100\_100\_0.15\_0.85$ and $85\_100\_0.15\_0.85$) for which BABC* has better values. Moreover, I2PLS has smaller standard deviations of its $f_{best}$ values ($f_{best}$ values often better than the compared results), suggesting that our algorithm is highly robust. 

Finally, the small $p$-$values$ ($<0.05$) of Table \ref{Summary_result} from the Wilcoxon signed-rank test (2.14e-4, 4.00e-6, 2.89e-5 and 1.43e-4) confirm that the results of our algorithm are significantly better than those of the compared results (best known in the literature, BABC, BABC* and gPSO).

\section{Analysis and Insights}
\label{Sec_Analysis}

In this section, we perform an analysis of the parameters and the ingredients of the algorithm to get useful insights about their impacts on its performance.

\subsection{Analysis of Parameters}
\label{Analysis of Parameters}

As shown in Table \ref{Parameter_Settings}, I2PLS requires four parameters: exploration depth $\lambda_{max}$ (Section  \ref{Algo_LOExplore}), 
neighborhood sampling probability $\rho$ (Section \ref{Descent}), tabu search depth  $\omega_{max}$ (Section \ref{Tabu}), perturbation strength $\eta$ (Sectiopn \ref{Escaping}). To analyze the sensibility and tuning of the parameters, we select 8 out of the 30 benchmark instances, i.e., 185\_200\_0.15\_0.85, 200\_185\_0.15\_0.85, 200\_200\_0.15\_0.85, 300\_285\_0.15\_0.85, 400\_385\_0.15\_0.85, 500\_485\_0.10\_0.75, 500\_485\_0.15\_0.85 and 500\_500\_0.15\_0.85. According to Table 6-9, the compared algorithms have a larger standard deviation for most of these instances than for other instances, implying that they are rather difficult to solve. We exclude the instances with 85 and 100 items since they can be solved exactly by the CPLEX and are thus too easy to be used for our analysis. 

\begin{figure}
\caption{Average of the best objective values ($f_{best}$) on the 8 instances obtained by executing I2PLS with different values of the four parameters.}
\label{Parameter_Analysis}

\begin{minipage}{0.5\textwidth} 
\includegraphics[width=2.8in]{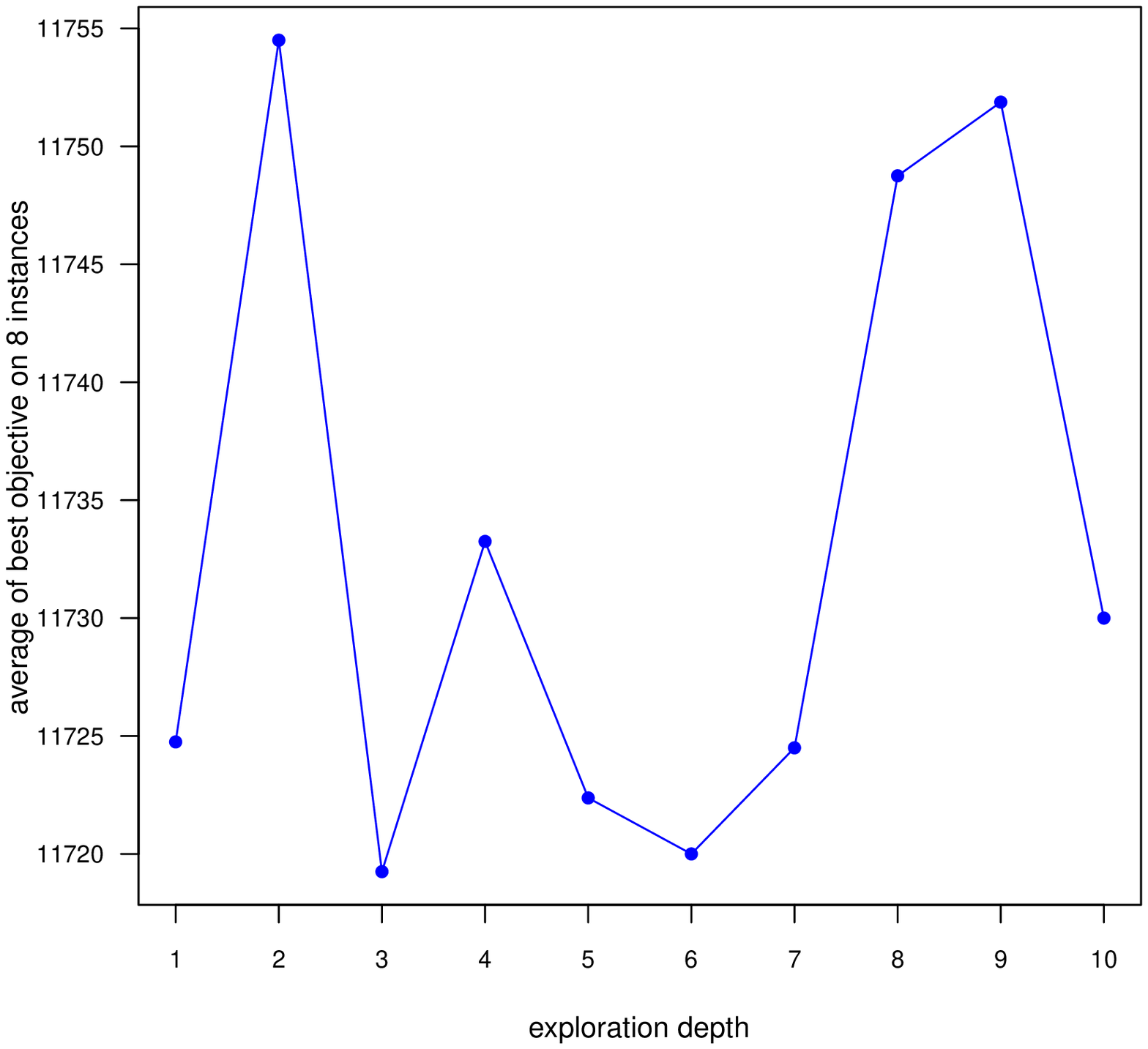}
\end{minipage}
\hfill 
\begin{minipage}{0.5\textwidth} 
\includegraphics[width=2.8in]{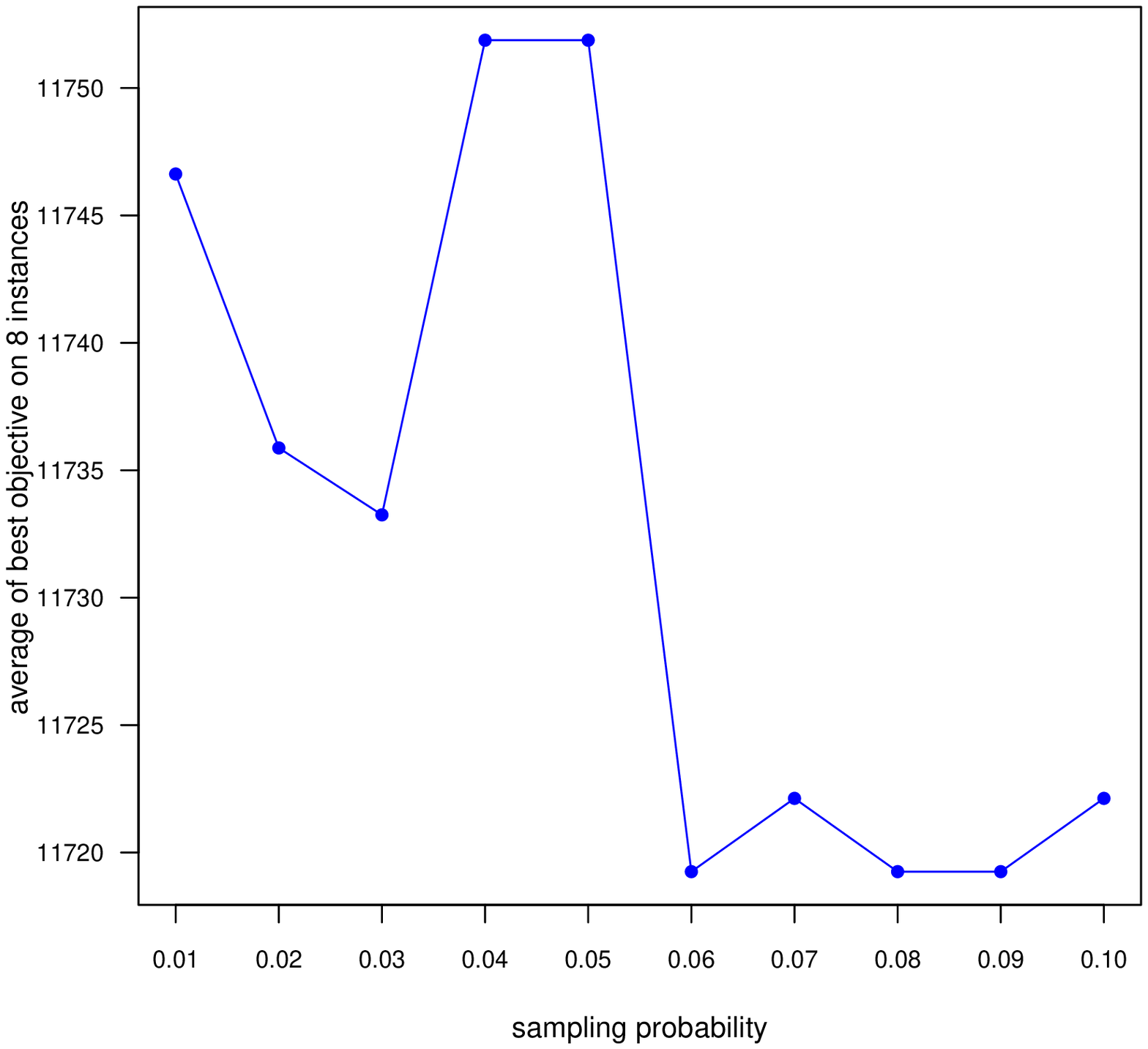}
\end{minipage}

\begin{minipage}{0.5\textwidth} 
\includegraphics[width=2.8in]{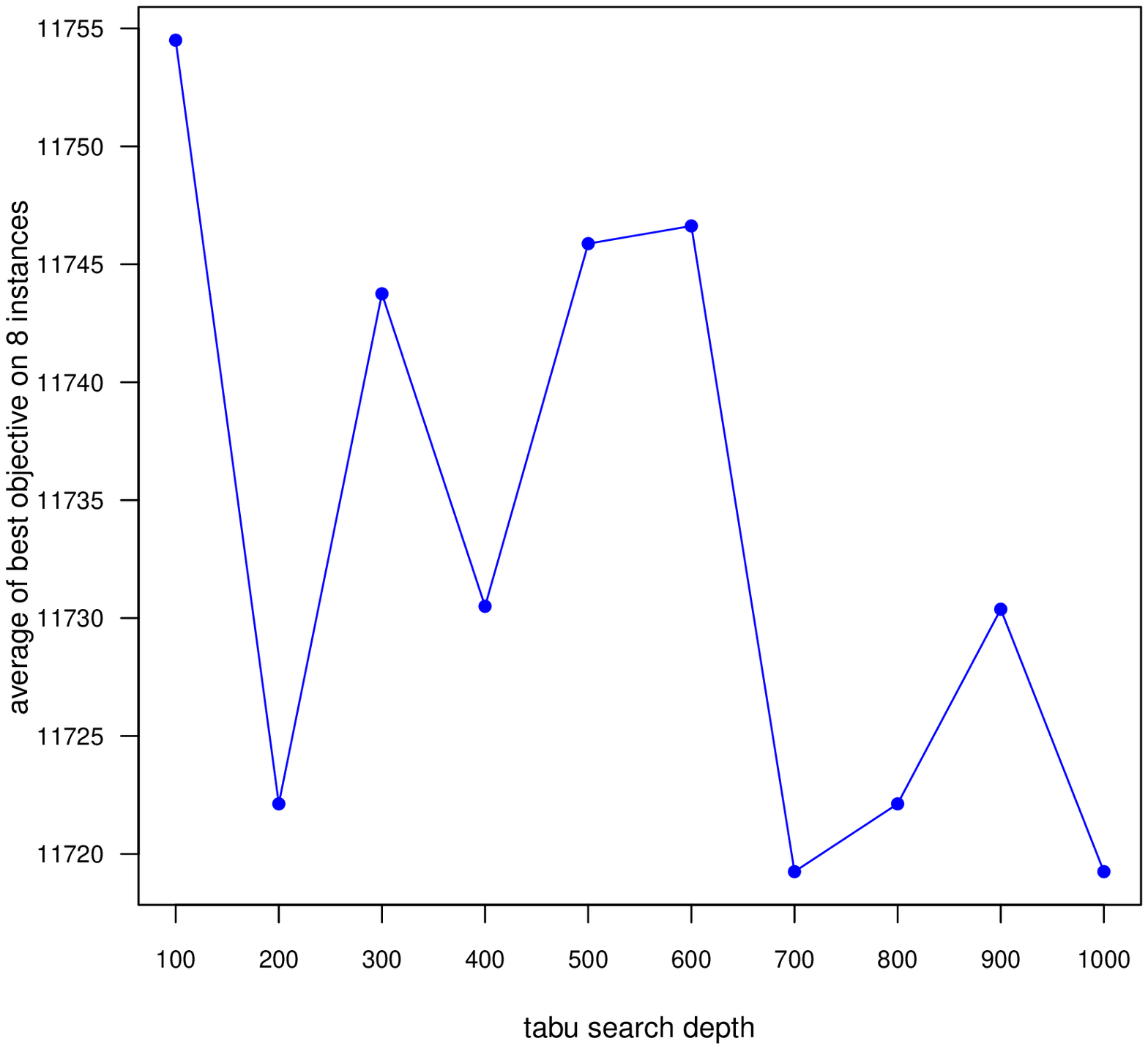}
\end{minipage}
\hfill 
\begin{minipage}{0.5\textwidth} 
\includegraphics[width=2.8in]{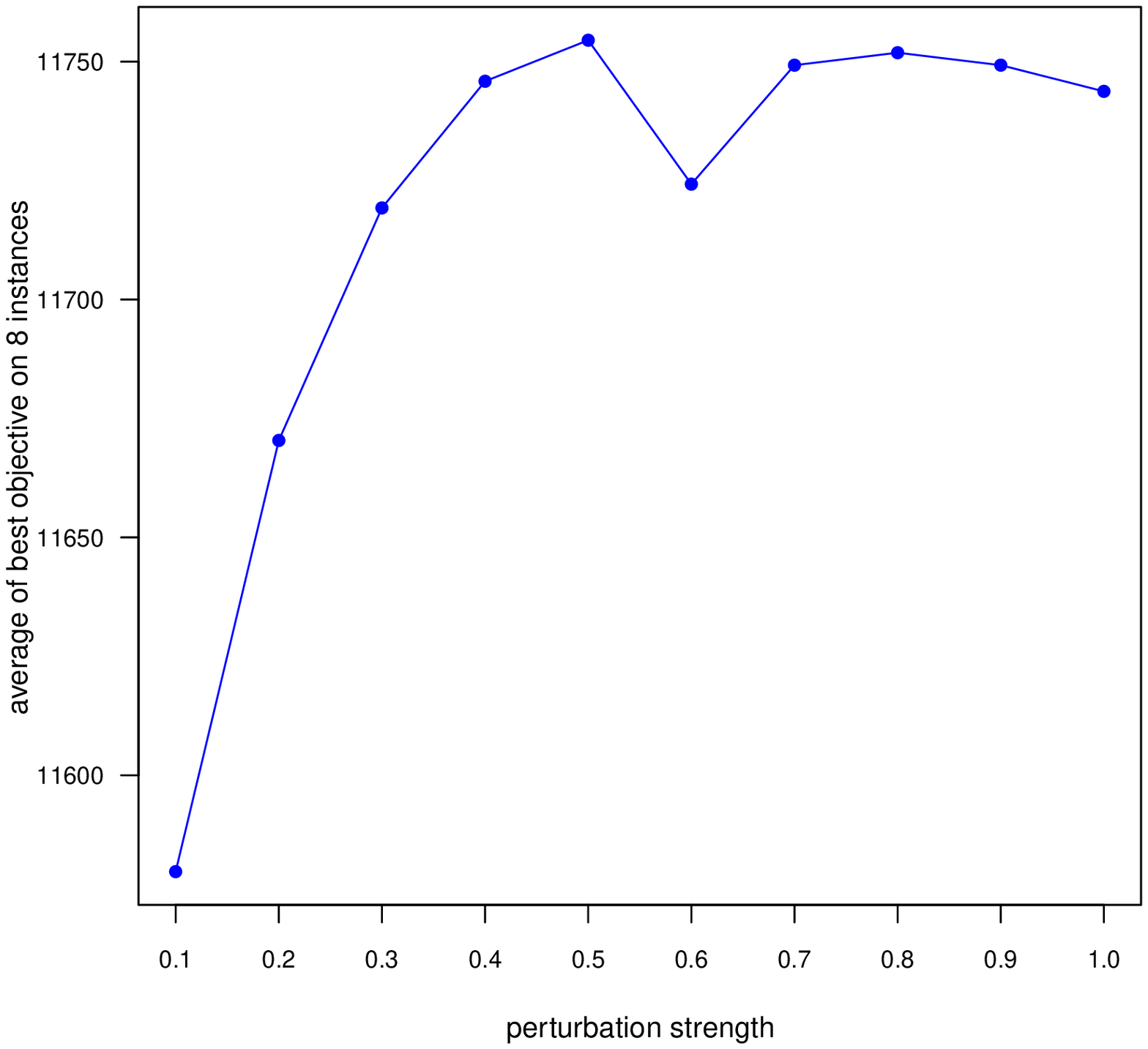}
\end{minipage}

\end{figure}

In this experiment, we studied each parameter independently by varying its value in a pre-determined range while fixing the other parameters to the default values shown in Table \ref{Parameter_Settings}. We then ran I2PLS with each parameter setting 30 times to solve each of the 8 selected instances with the same cut-off time as in Section \ref{Result}. Specifically, the exploration depth $\lambda_{max}$ takes its values in $\{1,2,\ldots,10\}$ with a step size of 1, the sampling probability $\rho$ varies from $0.01$ to $0.10$ with a step size of 0.01, the tabu search depth $\omega_{max}$ takes its values in $\{100,200,\ldots,1000\}$ with a step size of 100, and the perturbation strength $\eta$  varies from $0.1$ to 1 with a step size of 0.1. Figure \ref{Parameter_Analysis} shows the average of the best objective values ($f_{best}$) obtained by I2PLS with the four parameters on the 8 instances.  

Figure \ref{Parameter_Analysis} indicates I2PLS achieves better results when $\lambda_{max} = 2, \rho = 0.05$ (the $f_{avg}$ value is better when $\rho = 0.05$ than $\rho = 0.04$), $ \omega_{max} = 100, \eta = 0.5$, respectively. This justifies the adopted settings of parameters as shown in Table \ref{Parameter_Settings}. In addition, for each parameter, we used the non-parametric Friedman test to compare the $f_{best}$ values reached with each of the alternative parameter values. The resulting $p-value$ ($>$ 0.05) of the parameters $\lambda_{max}$ and $\omega_{max}$ show that the differences from alternative parameter settings are not statistically significant, implying that I2PLS is not sensitive to these two parameters.

\subsection{Effectiveness of the VND Search Strategy}
\label{Strategy1}

The VND procedure explores two neighborhoods $N_1$ and $N_2$ with a sampling probability $\rho$ applied to $N_2$. To investigate the impact of this sampling strategy, we performed an experiment by setting $\rho \in \{0.05, 0.0, 1.0\}$, where $\rho = 0.05$ is the adopted value as shown in Table \ref{Parameter_Settings}, $\rho = 0.0$ indicates that only the neighborhood $N_1$ is used during the descent search while $N_2$ is disabled, and $\rho = 1.0$ indicates that the entire neighborhoods $N_1$ and $N_2$ are explored. 

We denote these three VND variants by VND$_{0.05}$, VND$_{0.0}$ and VND$_{1.0}$ respectively. Recall that the VND procedure adopts the $best$-$improvement$ strategy at each iteration. However, it is interesting to observe the effect of adopting the $first$-$improvement$ strategy in $N_2$, So we included a fourth VND variant with the $first$-$improvement$ strategy and $\rho = 1.0$ (denoted as VND$_{1.0}^f$). We ran these four VND variants to solve the 30 benchmark instances under the condition of Section \ref{Result} and report the results in terms of $f_{best}$ in Table \ref{Summary_strategy1} (the best of the $f_{best}$ values in bold). The rows $\# Better$, $\# Equal$ and $\# Worse$ respectively indicate the number of instances for which VND$_{0.0}$, VND$_{1.0}$ and VND$_{1.0}^f$ attain a better, equal and worse result compared to the result obtained by VND$_{0.05}$ (which is the default strategy of I2PLS).
 
\renewcommand{\baselinestretch}{0.9}\large\normalsize
\begin{table}[!htp]\centering
	\caption{Influence of the VND search strategy on the performance of the I2PLS algorithm.} 
	\begin{scriptsize}
	\setlength{\tabcolsep}{2mm}{
	\begin{tabular}{llllll}
	\toprule[0.75pt]

Instance/Setting & VND$_{0.05}$ & VND$_{0.0}$ & VND$_{1.0}$ & VND$_{1.0}^f$ \\
\hline

100\_85\_0.10\_0.75 & 13283 & 13283 & 13283 & 13283 \\

100\_85\_0.15\_0.85 & 12479 & 12479 & 12479 & 12479  \\

200\_185\_0.10\_0.75 & 13521 & 13521 & 13521 & 13521 \\

200\_185\_0.15\_0.85 & 14215 & 14215 & 14215 & 14215 \\

300\_285\_0.10\_0.75 & 11563 & 11563 & 11563 & 11563 \\

300\_285\_0.15\_0.85 & \textbf{12607} & 12500 & 12332 & 12332 \\

400\_385\_0.10\_0.75 & 11484 & 11484 & 11484 & 11484 \\

400\_385\_0.15\_0.85 & 11209 & 11209 & 11209 & 11209 \\

500\_485\_0.10\_0.75 & \textbf{11771} & 11729 & 11746 & 11729 \\

500\_485\_0.15\_0.85 & \textbf{10238} & 10194 & 10194 & 10194 \\

100\_100\_0.10\_0.75 & 14044 & 14044 & 14044 & 14044 \\

100\_100\_0.15\_0.75 & 13508 & 13508 & 12238 & 13508 \\

200\_200\_0.10\_0.75 & 12522 & 12522 & 12522 & 12522 \\

200\_200\_0.15\_0.85 & 12317 & 12317 & 12317 & 12317 \\

300\_300\_0.10\_0.75 & 12817 & 12817 & 12817 & 12817 \\

300\_300\_0.15\_0.85 & 11585 & 11585 & 11502 & 11585 \\

400\_400\_0.10\_0.75 & 11665 & 11665 & 11665 & 11665 \\

400\_400\_0.15\_0.85 & 11325 & 11325 & 11325 & 11325 \\

500\_500\_0.10\_0.75 & 11249 & 11249 & 11249 & 11249 \\

500\_500\_0.15\_0.85 & 10381 & 10381 & 10381 & 10381 \\

85\_100\_0.10\_0.75 & 12045 & 12045 & 12045 & 12045 \\

85\_100\_0.15\_0.85 & 12369 & 12369 & 12369 & 12369 \\

185\_200\_0.10\_0.75 & 13696 & 13696 & 13696 & 13696 \\

185\_200\_0.15\_0.85 & 11298 & 11298 & 11298 & 11298 \\

285\_300\_0.10\_0.75 & 11568 & 11568 & 11568 & 11568 \\

285\_300\_0.15\_0.85 & 11802 & 11802 & 11802 & 11802 \\

385\_400\_0.10\_0.75 & 10600 & 10600 & 10600 & 10600 \\

385\_400\_0.15\_0.85 & 10506 & 10506 & 10506 & 10506 \\

485\_500\_0.10\_0.75 & 11321 & 11321 & 11321 & 11321 \\

485\_500\_0.15\_0.85 & 10220 & 10220 & 10220 & 10208 \\

\hline
\# Better & - & 0 & 0 & 0 \\

\# Equal  & - & 27 & 25 & 26 \\
 
\# Worse  & - & 3 & 5 & 4 \\

\bottomrule[0.75pt]
	\end{tabular}}

\label{Summary_strategy1}
\end{scriptsize}
\end{table} 
\renewcommand{\baselinestretch}{1}\large\normalsize

Table \ref{Summary_strategy1} shows that VND$_{0.05}$ performs the best with the setting $\rho=0.05$. Compared to VND$_{0.05}$, VND$_{0.0}$ obtains worse results on 3 instances, and equal results on the other 27 instances. VND$_{1.0}$ reaches the same results as VND$_{0.05}$ on 25 instances, and worse results on 5 instances. VND$_{1.0^f}$ obtains worse results on 4 instances, and equal results on the other 26 instances. Moreover, we observe that when exploring the whole neighborhood $N_2$, neither the $best$-$improvement$ strategy nor the $first$-$improvement$ strategy performs well. This can be explained by the fact that given the large size of $N_2$, a thorough examination of this neighborhood becomes very expensive. Within the cut-off time, the VND search cannot perform many iterations, decreasing its chance of encountering high-quality solutions. Finally, the $p-value$ of 4.18e-2 from the Friedman test indicates a significant difference among the compared VND strategies. This implies that the adopted VND strategy and sampling technique of the I2PLS algorithm are relevant for its performance.

\subsection{Effectiveness of the Frequency-Based Local Optima Escaping Strategy}
\label{Strategy2}

The frequency-based local optima escaping strategy of I2PLS perturbs the locally best solution $S_b=(A,\bar{A})$ by replacing the first $\eta \times |A|$ (in I2PLS, $\eta$ is set to $0.5$) least frequently moved items of $A$ with items that are randomly chosen from $\bar{A}$. In this experiment, we compared I2PLS against two variants with alternative perturbation strategies. In the first variant (denoted by I2PLS$_{random}$), we replace $0.5 \times |A|$ items randomly selected items of $A$ while in the second variant (denoted by I2PLS$_{strong}$) and we perform a very strong perturbation by replacing all the items of $A$ with items of $\bar{A}$ (i.e., setting $\eta$ to 1). We ran I2PLS, I2PLS$_{random}$ and I2PLS$_{strong}$ 30 times to solve each of the 30 benchmark instances. The computational results of this experiment are shown in Table \ref{Summary_strategy2} where in addition to the best $f_{best}$ values of each compared algorithm (the best of the $f_{best}$ values in bold), the last three rows indicate the number of instances for which I2PLS$_{random}$ and I2PLS$_{strong}$ has a better, equal and worse result compared to that of I2PLS.

\renewcommand{\baselinestretch}{0.9}\large\normalsize
\begin{table}[!htp]\centering
	\scriptsize
	\caption{Impact of the frequency-based local optima escaping strategy on the performance of the I2PLS algorithm.} 
	\centering
	\setlength{\tabcolsep}{2mm}{
	\begin{tabular}{lllll}
	\toprule[0.75pt]

Instance/Setting & I2PLS & I2PLS$_{random}$ & I2PLS$_{strong}$ \\
\hline

100\_85\_0.10\_0.75 & 13283 & 13283 & 13283 \\

100\_85\_0.15\_0.85 & 12479 & 12479 & 12479 \\

200\_185\_0.10\_0.75 & 13521 & 13521 & 13521 \\

200\_185\_0.15\_0.85 & 14215 & 14215 & 14215 \\

300\_285\_0.10\_0.75 & 11563 & 11563 & 11563 \\

300\_285\_0.15\_0.85 & 12607 & 12607 & 12607 \\

400\_385\_0.10\_0.75 & 11484 & 11484 & 11484 \\

400\_385\_0.15\_0.85 & 11209 & 11209 & 11209 \\

500\_485\_0.10\_0.75 & \textbf{11771} & 11729 & 11729 \\

500\_485\_0.15\_0.85 & \textbf{10238} & 10194 & 10194 \\

100\_100\_0.10\_0.75 & 14044 & 14044 & 14044 \\

100\_100\_0.15\_0.75 & 13508 & 13508 & 13508 \\

200\_200\_0.10\_0.75 & 12522 & 12522 & 12522 \\

200\_200\_0.15\_0.85 & 12317 & 12317 & 12317 \\

300\_300\_0.10\_0.75 & 12817 & 12817 & 12817 \\

300\_300\_0.15\_0.85 & 11585 & 11585 & 11585 \\

400\_400\_0.10\_0.75 & 11665 & 11665 & 11665 \\

400\_400\_0.15\_0.85 & 11325 & 11325 & 11325 \\

500\_500\_0.10\_0.75 & 11249 & 11249 & 11249 \\

500\_500\_0.15\_0.85 & 10381 & 10381 & 10381 \\

85\_100\_0.10\_0.75 & 12045 & 12045 & 12045 \\

85\_100\_0.15\_0.85 & 12369 & 12369 & 12369 \\

185\_200\_0.10\_0.75 & 13696 & 13696 & 13696 \\

185\_200\_0.15\_0.85 & 11298 & 11298 & 11298 \\

285\_300\_0.10\_0.75 & 11568 & 11568 & 11568 \\

285\_300\_0.15\_0.85 & 11802 & 11802 & 11802 \\

385\_400\_0.10\_0.75 & 10600 & 10600 & 10600 \\

385\_400\_0.15\_0.85 & 10506 & 10506 & 10506 \\

485\_500\_0.10\_0.75 & 11321 & 11321 & 11321 \\

485\_500\_0.15\_0.85 & 10220 & 10220 & 10220 \\

\hline
\# Better & - & 0 & 0  \\

\# Equal  & - & 28 & 28 \\

\# Worse  & - & 2 & 2  \\

\bottomrule[0.75pt]
	\end{tabular}}

\label{Summary_strategy2}

\end{table} 
\renewcommand{\baselinestretch}{1.0}\large\normalsize

Table \ref{Summary_strategy2} shows that I2PLS with its frequency-based local optima escaping strategy performs slightly better than the two variants with alternative perturbation strategies. Indeed, even if the compared strategies lead to equal results for 28 instances, I2PLS achieves a better result on two of the most difficult instances (500\_485\_0.10\_0.75 and 500\_485\_0.15\_0.85). This experiment tends to indicate that the frequency-based local optima escaping strategy is particularly helpful for solving difficult instances. The $p-value$ of 1.35e-1 from the Friedman test indicates that the compared strategies differ only marginally.

\section{Conclusions}
\label{Conclusions}

The set-union knapsack problem (SUKP) studied in this work is a generalization of the conventional 0-1 knapsack problem with a variety of practical applications. Existing solution methods are mainly based on swarm optimization. This work introduces the first local search approach for solving the SUKP that directly operates in the discrete search space. The proposed algorithm combines a local optima exploration phase and a local optima escaping phase based on frequency information within the iterated local search framework.

The proposed algorithm has been tested on three sets of 30 benchmark instances commonly tested in the literature and showed a high competitive performance compared to the state-of-the-art SUKP algorithms. Specifically, our algorithm has improved on the best-known results (new lower bounds) for 18 out of the 30 benchmark instances, while matching the best-known results for the remaining 12 instances. Moreover, we has investigated for the first time the interest of the general mixed integer linear programming solver CPLEX for solving the SUKP, showing that the optimal solutions can be reached only for 6 small instances. Furthermore, we have analyzed the impacts of parameters and the main components of the algorithm on its performance.

This work can be further improved. First, even if the algorithm uses the filtering mechanism and the sampling technique to reduce the neighborhoods, evaluating a given neighbor solution remains time-comsuming. To speed up the search process, it is useful to seek streamlining techniques to reduce the complexity of neighborhood evaluation. Second, considering the potential strong correlations of constituent elements between different items, a hybrid approach combining  local search and population-based search could be helpful to break search barriers and traps. Finally, the SUKP belongs to the large family of knapsack problems, it would be interesting to investigate whether proven techniques and strategies designed for related knapsack problems remain useful for solving the SUKP.

\section*{Acknowledgments}
%\bibitem{he2018novel}
We would like to thank Prof. Yichao He and his co-authors for making the benchmark instances tested in \cite{he2018novel} available and sharing the codes of their BABC algorithm with us.

\bibliographystyle{plain}

\begin{appendix}

\section{Appendix: 0/1 linear programming model for the SUKP}
\label{0/1 linear programming model for the SUKP}

Based on the mathematical model of \cite{he2018novel}, we introduced a modified 0/1 linear programming model of the SUKP that is suitable for the general ILP solver CPLEX. For an arbitrary non-empty item set $S \subset V$, let $y_i$ $(i=1,\ldots,m)$ be a binary variable such that $y_i = 1$ if item $i$ is selected (i.e., $i \in S$), and $y_i = 0$ otherwise (i.e., $i \notin S$). Let $R$ be a $m \times n$ binary relation matrix such that $R_{ij} = 1$ if element $j$ belongs to item $i$, and $R_{ij} = 0$ otherwise. Furthermore, for each element $j \ (j=1,\ldots,n)$, define $L_j = \sum\limits_{i=1}^m y_i R_{ij}$ that counts the number of appearances of element $j$ in the items of $S$. Let $x_j$ be a binary variable such that $x_j = 1$ if $L_j > 0$, and $x_j = 0$ otherwise, that is, $x_j$ indicates whether element $j$ is involved in calculating the total weight of $S$. Then the SUKP can be formulated as the following integer linear program.

\begin{equation}\label{FMAX}
(SUKP) \quad  \mathrm{Maximize} \quad \sum\limits_{i = 1}^m p_iy_i
\end{equation}
\begin{equation}\label{constraint2}
\mathrm{s.t.} \quad  \sum\limits_{j = 1}^n w_jx_j \leq C
\end{equation}	
\begin{equation}\label{constraint3}
x_j = 
   \begin{cases}
		1, & if \ L_j > 0; \\
		0, & \text{otherwise}.
		\end{cases}
\end{equation}
\begin{equation}
L_j = \sum\limits_{i=1}^m y_i R_{ij}, j=1,\ldots,n
\end{equation}
\begin{equation}\label{constraint4}
%\quad \quad \quad x_j = \begin{cases}
		y_i \in \{0,1\}, i=1,\ldots,m
%\end{cases}
\end{equation}

The results of CPLEX reported in Section \ref{Result} are based on this formulation.

\end{appendix}
\end{document}